\documentclass{report}
\usepackage{graphicx}
\usepackage{ohm-ngc}
\usepackage{latexsym}

\begin{document}
\title{Interactive Restless Multi-armed Bandit Game
and Swarm Intelligence Effect}
\author{Shunsuke Yoshida}{Kitasato University\\ 1-15-1 Kitasato, Sagamihara, Kanagawa 252-0373 JAPAN}
\author{Masato Hisakado}{Financial Services Agency\\ 3-2-1 Kasumigaseki, Chiyoda-ku, Tokyo 100-8967 JAPAN}
\author{Shintaro Mori}{Kitasato University\\ 1-15-1 Kitasato, Sagamihara, Kanagawa 252-0373 JAPAN}
\E-mail{shintaro.mori@gmail.com}
\date{9 September 2014}
\begin{abstract}
 We obtain the conditions for the emergence of the swarm intelligence effect in
 an interactive game of restless multi-armed bandit (rMAB).
 A player competes with multiple agents.
 Each bandit has a payoff that
 changes with a probability $p_{c}$ per round.
 The agents and player choose one of three options: (1)
 Exploit (a good bandit), (2)
 Innovate (asocial learning for a good bandit among $n_{I}$ randomly chosen
 bandits), and
 (3) Observe (social learning for a good bandit).
 Each agent has two parameters $(c,p_{obs})$ to specify the decision: (i)
 $c$, the threshold value for Exploit, and
 (ii) $p_{obs}$, the probability for Observe in learning.
 The parameters $(c,p_{obs})$ are uniformly distributed.
 We determine the optimal strategies for the player
 using complete knowledge about the rMAB.
 We show whether or not social or asocial
 learning is more optimal in the $(p_{c},n_{I})$ space and define the swarm intelligence effect.
 We conduct a laboratory experiment (67 subjects) and observe
 the swarm intelligence effect only if $(p_{c},n_{I})$ are chosen so that
 social learning is far more optimal than asocial learning.
\end{abstract}

\section{Introduction}
The trade-off between the exploitation of good choices and the exploration
 of unknown but potentially more profitable choices is a well-known
problem \cite{Kam:2003,Ren:2010,Gue:2014}. A multi-armed bandit (MAB)
provides the most typical environment
for studying this trade-off.
It is defined by sequential decision making among multiple
choices that are associated with a payoff. The MAB
problem involves the maximization of
the total reward for a given period or budget.
In a variety of circumstances, exact or approximated
optimal strategies have been proposed
\cite{Ber:1985,Lai:1985,Sut:1998,Aue:2002,Whi:2012}.

Recently, the MAB has also provided a good environment for the
trade-off between social and asocial learning \cite{Ren:2010}.
Here, social learning is learning through
observation or interaction with other individuals,
and asocial learning is individual learning
\cite{Lal:2004,Ren:2010,Kam:2003,Toy:2014}.
The advantage of social learning is its cost
compared with asocial learning.
The disadvantage is its error-prone nature, as
the information obtained by social learning
might be outdated or inappropriate.
In order to clarify the optimal strategy in the environment with 
 the two trade-offs, Rendell et al. held
 a computer tournament using a restless multi-armed bandit (rMAB)
 \cite{Ren:2010}.
 Here, restless means that the payoff of each bandit changes over time.
 There are 100 bandits in an rMAB, and each
 bandit has a distinct payoff independently drawn from an
 exponential distribution. The probability that a payoff changes
 per round is $p_{c}$. An agent has three options for each round:
 Innovate, Observe, and Exploit.
 Innovate and Observe correspond to asocial and
 social learning, respectively.
 For Innovate, an agent obtains the payoff information of one randomly
 chosen bandit. For Observe, an agent obtains the payoff information
 of $n_{O}$ randomly chosen bandits
 that were exploited by the agents during the previous round.
 Compared to the information obtained by Innovate,
 that obtained by Observe is older by one round.
 For Exploit, an agent chooses a bandit that
 he has already explored by Innovate or Observe and obtains a payoff.
 In an rMAB environment, it is extremely difficult
 for agents to optimize their choices \cite{Pap:1999,Ren:2010}.
 The outcome of the tournament was that the winning strategies
 relied heavily on social learning. This contradicted
 previous studies in which
 the optimal strategy is a mixed one
 that relies on some combination
 of social and asocial learning.
 In the tournament, the cost for Observe was not very low, as approximately
 50\% of the choices of Observe returns information that 
 the agents already knew. 
 The results of the tournament imply
 the inadvertent filtering of information 
 when an agent chooses Observe, as
 the agents choose the best bandit during Exploit.

 In this paper, we discuss whether social or asocial
 learning is optimal in an
 rMAB, where a player competes with many agents.
 We answer to the question why social learning is so adaptive in Rendell's
 tournament. We suppose that the cost of Innovate becomes
 higher than that of Observe in the tournament.
 In order to reduce the cost of Innovate, we control
 the exploration range $n_{I}$ for Innovate, and agents
 obtain the best information about 
 the bandits among $n_{I}$ randomly chosen bandits.
 An rMAB is characterized by two parameters, $p_{c}$ and $n_{I}$.
 We compare the average payoffs of the optimal
 strategies when only Innovate, only Observe, and both are
 available for learning using the complete knowledge
 of an rMAB and the information of the bandits exploited by agents.
 We determine the region in which each type of learning is
 optimal in the $(n_{I},p_{c})$ plane and show that
 Observe is more adaptive than Innovate for $n_{I}=1$.
 We define the swarm intelligence effect as the
 increase in the average payoff compared with
 the payoffs of the optimal strategies where only asocial
 learning is available.
 We have conducted a laboratory experiment where
 67 human subjects competed with multiple agents in an rMAB.
 If the parameters are chosen in the region where social learning
 is far more optimal than asocial learning, we observe the swarm
 intelligence effect.

\section{Restless multi-armed bandit interactive game}
\begin{figure}[tb]
  \begin{center}
  \begin{tabular}{cc}
  \includegraphics[width=60mm]{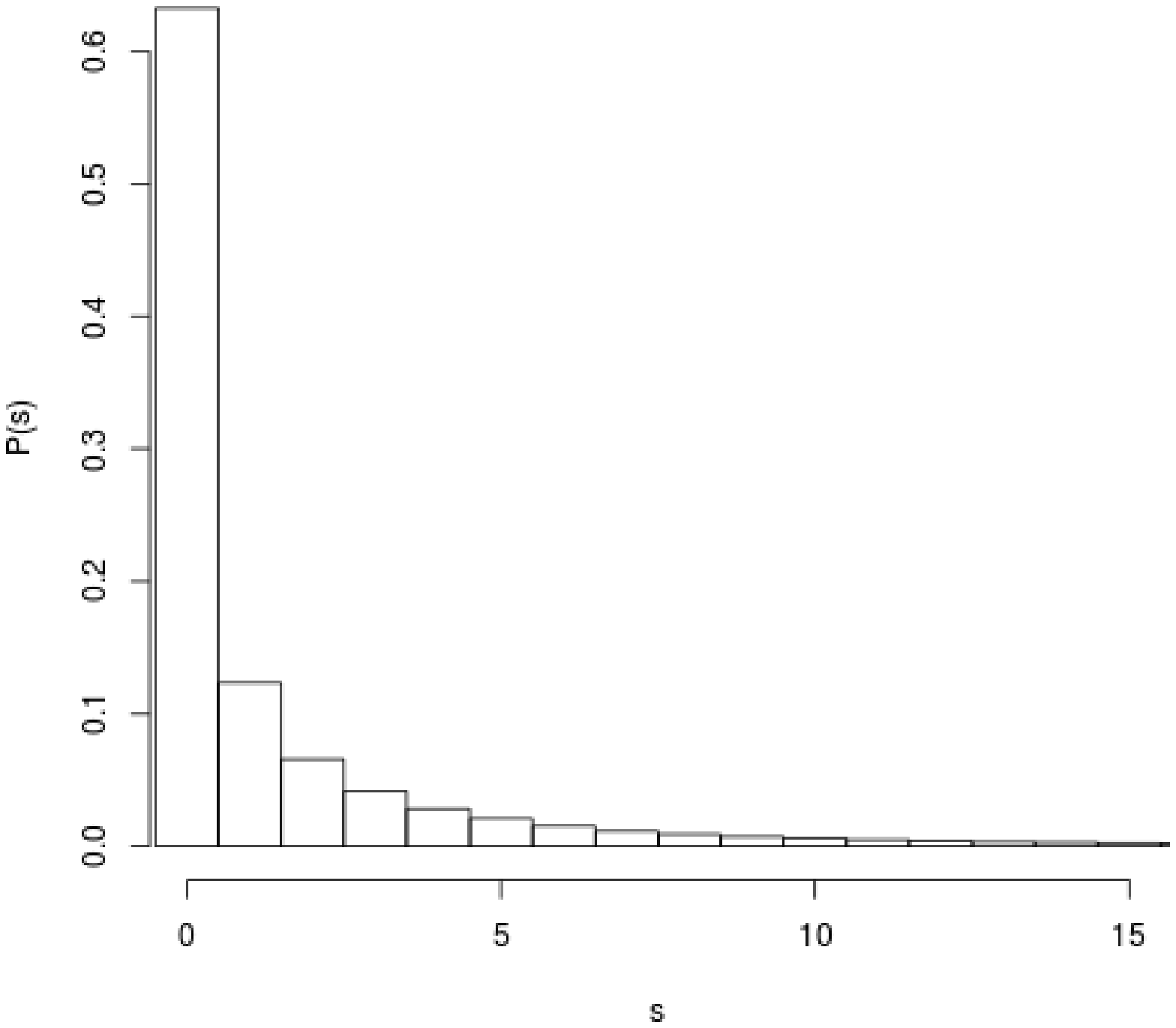} \
  \includegraphics[width=70mm]{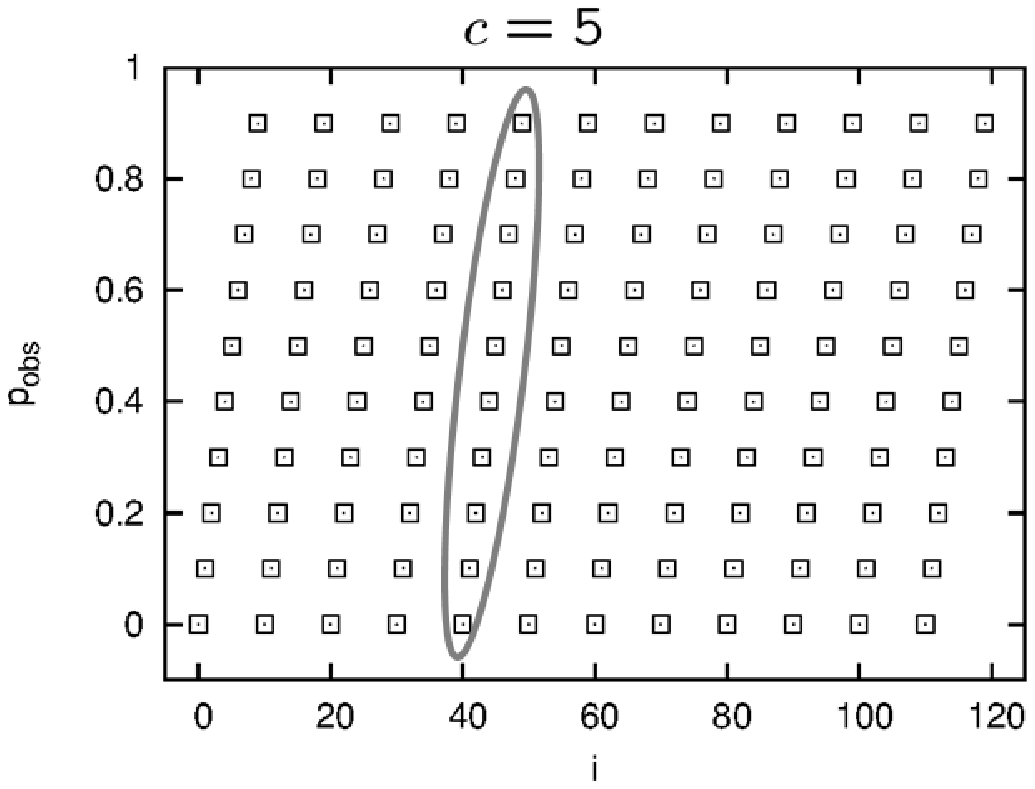}
  \end{tabular}
  \end{center}
  \caption{Left: Plot of $P(s)$. The expected value of $s$ is 
E$(s)\simeq  1.68$.
Right: Parameter assignment for agent $i\in \{1,2,\cdots,120\}$.
$p_{obs}(i)=0.1\times (i\%10) \in \{0.0,0.1,\cdots,0.9\} $.
$c(i)=i/10+1\in \{1,2,\cdots,12\}$.
}
\label{fig:game}
\end{figure}

 An interactive rMAB game is a game in which
 a player competes with 120 agents using an rMAB.
 The player aims to
 maximize the total payoff over 103 rounds
 and obtain a high ranking among all entrants.
 Below, we term the population of all agents and a player as all entrants.
 The rMAB has $N=100$ bandits, and we label them as
 $n \in \{1,2,\cdots,N=100\}$.
 Bandit $n$ has a distinct payoff $s(n)$, and we
 term the $(n,s(n))$ pair as bandit information.
 $s(n)$ is an integer drawn at random
 from an exponential distribution ($\lambda=1$; values
 were squared and
 rounded to give integers mostly falling in the range 
 of 0--10 \cite{Ren:2010}).
 We denote the probability function
 for $s(n)$ as $\mbox{Pr}(s(n)=s)=P(s)$ 
 (left figure in Figure \ref{fig:game}).
 We write the expected
 value of $s(n)$ as E$(S(n))$, and it is approximately 1.68.
 The payoff of each bandit changes independently between rounds
 with a probability $p_{c}$, with
 new payoff drawn at random from the same distribution.

 Every entrant has his own repertoire and can store at most three pieces of
 bandit information. The bandit information has a time stamp when
 the entrant obtains it.
 The time stamp is updated when the entrant obtains new
 bandit information about the bandit.
 When an entrant obtains more than three pieces of bandit
 information, the one with the oldest time stamp is
 erased from the repertoire.

 There are three possible moves for the entrants: 
Innovate, Observe, and Exploit.
 Innovate and Observe are learning
 processes to obtain bandit information. Exploit is the exploitation
 process that obtains some payoff.

\begin{itemize}
\item Innovate is individual learning, and an entrant obtains bandit
 information. $n_{I}$ bandits are chosen at random among $N=10^{2}$ bandits, and
 the bandit information with the maximum payoff is provided to the entrant.
 If there are several bandits with the same maximum payoff, one of them
 is chosen at random.

\item Observe is social learning, and an entrant
 obtains the bandit information exploited by an 
 agent during the previous round.
 If there are many agents who exploited a bandit, an agent
 is randomly chosen among them, and its bandit information 
 is provided to the entrant.
 If there are no such agents, no bandit information is provided.
 The information obtained by Observe is one round older 
 than that obtained by Innovate.

\item Exploit is the exploitation of a bandit. An entrant chooses a bandit
 from his repertoire and exploits the bandit. Even if the bandit information
 is $(n,s(n))$, as the information changes with a probability $p_{c}$
 per round, he does not necessarily receive the payoff $s(n)$.
\end{itemize}

The repertoire is updated after a move.
For Innovate, the bandit information with the maximum payoff $s_{I}$ among
 $n_{I}$ randomly chosen bandits is provided to the entrant. 
 We denote the distribution function of $s_{I}$ as 
 $P_{I}(s)=\mbox{Pr}(s_{I}=s)$. Intuitively, $s_{I}$ is chosen
 in the region of upper probability $1/n_{I}$ of $P(s)$.
We denote the expectation value of $s_{I}$ as E$(s_{I})$.
If $n_{I}>1$, E$(s_{I})>\mbox{E}(s)$ holds. For example, E$(s_{I})
\simeq 9.63$ for
$n_{I}=10$. By controlling $n_{I}$, we can change the cost
of Innovate.

\subsection{Agent strategy}
We explain the strategy of the agents.
The most important factor in the performance of the
strategies in Rendell's tournament was
the proportion of Observe in learning \cite{Ren:2010}.
The high performance of Observe originated from the inadvertent
filtering of bandit information, as the agents exploited
the best bandit in their repertoires. If the agents choose at random,
 Observe does not provide good bandit information.
 We take these facts into account and introduce a simple
 strategy for the agents with two parameters $c$ and $p_{obs}$.

\begin{itemize}
\item $c$: every agent has a threshold value $c$. If there is no bandit
 in one's repertoire whose
 payoff is greater than $c$, the agent will learn by Innovate or Observe.

\item $p_{obs}$: an agent chooses Observe with a probability
$p_{obs}$ when he learns.
\end{itemize}

We label 120 agents as $i\in \{1,2,\cdots,I=120\}$. Agent $i$ has
the parameters $(c(i),p_{obs}(i))$. $c(i)$ is given as the quotient $i/10$
plus one.
$p_{obs}(i)$ is the remainder of $i\%10$ multiplied by 0.1.
The assignment of $(c,p_{obs})$ to agent $i$ is represented in the right
 figure in Figure \ref{fig:game}.

\subsection{Game environment}
A player participated in a game and
competed with $N$ agents. However, the game did not
advance on a real-time basis. Agents had already 
participated in the game for 1000 rounds.
 When a player participated in the game, 103
sequential rounds were randomly chosen from the 1000 rounds, and
he competed with agents for 103 rounds.
We denote the round by $t\in \{-2,-1,0,1,2,\cdots,T=100\}$.
The scores of the player and agents were set to zero.
The agents had already
stored at most three pieces of bandit information in their repertoires.
The player had three rounds to learn the rMAB.
He could choose Innovate or Observe for three rounds and
stored at most three pieces of bandit information in his repertoire.
After three rounds, the rMAB game started.
As the agents had already finished the game, they could not observe
 the information of the player. On the other hand, the player
 could observe the information of the agents.

\begin{figure}[t]
\begin{center}
\includegraphics[width=70mm]{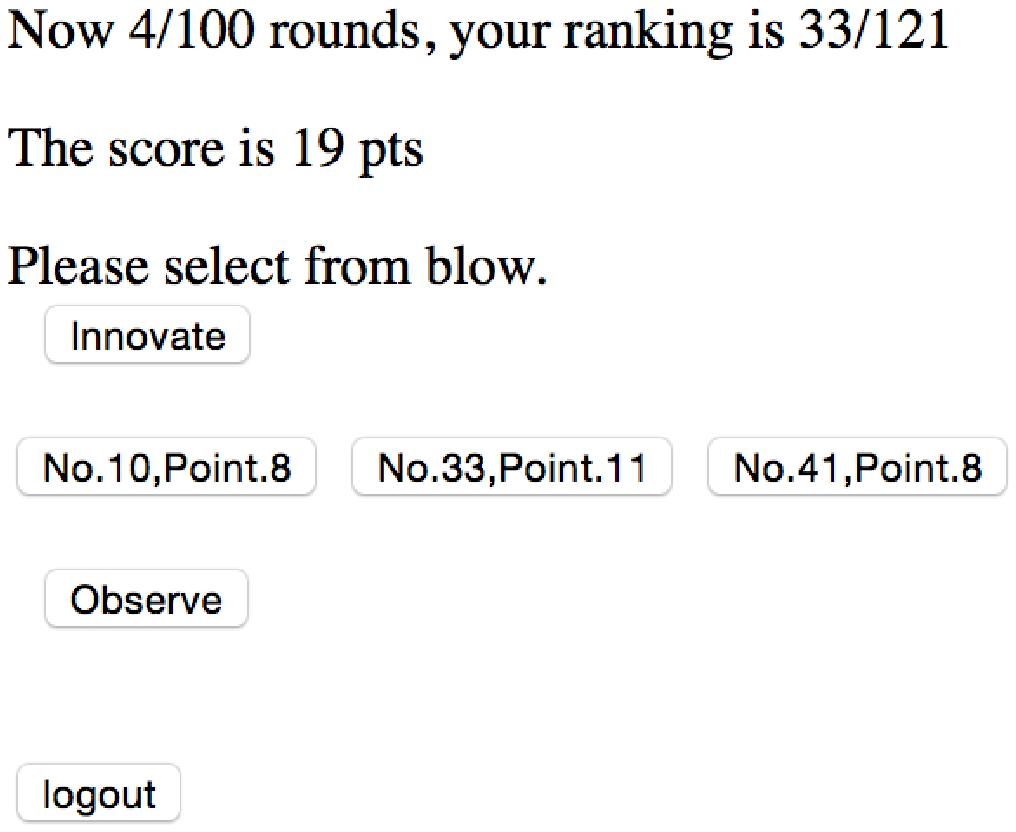}
\caption{The interactive rMAB game
online interface. A human player is presented with the
present round $t/100$, his ranking among 121 entrants (one player and
120 agents), and his repertoire. He must choose one among Innovate, Exploit a bandit, and Observe. In his repertoire, only $(n,s(n))$ is shown.
The bandit information from left to right indicates the newest to oldest information, respectively.
}
\label{fig:gameSc}
\end{center}
\end{figure}

The game environment was constructed as a website.
The information of the agents for 1000 rounds was
stored in a database of the website.
The player used a tablet (7 inch) and participated in the game through a
web browser.
The player had to learn for three rounds and stored at most
three pieces of bandit information in his repertoire. Afterwards, the
game started. Figure \ref{fig:gameSc} shows the interface of
the rMAB game. For the present round $t$, the ranking and score are
shown on the screen. The player had to choose an action
among Innovate, Exploit, and Observe. For Exploit, the player had to
choose which bandit he would exploit in his repertoire.
Then, the payoff and new ranking were shown on the screen,
and the game proceeded to the next round.

For the parameters $(n_{I},p_{c})$ of the rMAB,
we adopted the next four combinations. We call the combinations
A, B, C, and D.
\begin{itemize}
\item[A:]$(n_{I},p_{c})=(1,0.1)$. $p_{c}$ is small and 
the change in the payoff of a bandit is slow.
As $n_{I}=1$, $\mbox{E}(s_{I})=\mbox{E}(s)$, and it is difficult to find a
bandit with high payoff with Innovate.

\item[B:]$(n_{I},p_{c})=(10,0.1)$. $p_{c}$ is small, as in A. As $n_{I}=10$,
$\mbox{E}(s_{I})\simeq 9.63$ is large, and good bandit
information can be obtained with Innovate.

\item[C:]$(n_{I},p_{c})=(1,0.2)$. $p_{c}$ is large, and the bandit information changes
frequently. As $n_{I}=1$, it is difficult to obtain good bandit information with Innovate.

\item[D:]$(n_{I},p_{c})=(10,0.2)$. $p_{c}$ is large, 
as in C. As $n_{I}=10$, good bandit information 
can be obtained with Innovate.

\end{itemize}

\subsection{Experimental procedure}
The experiment reported here were conducted at the Information
Science room at Kitasato University. The subjects included
students from the university, mainly from the School of Science.
The number of subjects $S$ was 67.
Each subject participated in the game at most four times.

The subjects entered a room and sat down on a chair.
After listening to a brief explanation about the
experiment and reward, they signed a consent document for
participation in the experiment. Afterwards, they
logged into the experiment website using the IDs written 
on the consent document.
The game environment was chosen among the four cases A, B, C, and D,
and they started their games. After $100+3$ rounds, the game ended.
The subjects logged into the website again to participate in a new game.
Within the allotted time of approximately 40 min, most subjects 
participated in the game at least three times.
Subjects were paid upon being released from the experiment.

There were slight differences in the experimental setup and
rewards among the subjects.
For the first 21 subjects (July 2014), there was no participation fee.
The reward was completely determined by the number of times
that they entered the Top 20 among the $120+1$ entrants in each game.
Their rewards were a prepaid card of 300 yen (approximately \$2.50) 
for each placement within the Top 20.
The subject could choose the game environment at the start of the game.
They could choose each environment at most once,
and the average number of subjects in each environment is approximately 19.
They did not know the parameters of each environment.
For the last 46 subjects (December 2014) there was a 1050 yen (approximately \$9)
participation fee in addition to the performance-related reward.
The reason for the change in the reward is to recruit more subjects.
They were asked to play the game at least three times during 
the allotted time.
The game environment was randomly chosen by the experimental program.
The average number of subjects in each environment is approximately 37.
A total of 67 subjects participated in the experiment, and we gathered
data from approximately 56 subjects for each game environment.

\section{Optimal strategy and swarm intelligence effect}

 We estimate the expected payoff of the optimal strategies for the player
 in the rMAB game. Here, optimal means to maximize the expected total payoff in
 a total of $100+3$ rounds. For the first three rounds ($t\in \{-2,-1,0\}$), the player
 could choose Innovate or Observe. After that, he could choose all three options.
 The optimal choice for round $t$ is defined as the choice that maximizes the expected
 payoff obtained during the remaining $T-t$ rounds.

 We assume that the player has the complete knowledge about the rMAB game.
 More concretely, he knows $p_{c}$, $\mbox{E}(s)$, and $\mbox{E}(s_{I})$ about the rMAB.
 Furthermore, he knows the bandit information exploited during the previous round.
 We denote the average value of the payoff of the exploited bandit at round $t-1$ as
 $\overline{O}(t)$.
 If the player chooses Observe for round $t$, the expected value of the payoff of the obtained bandit
 information is $\overline{O}(t)$. $\overline{O}(t)$ depends on the agents' choices in the background.
 It is usually the most difficult quantity to estimate for the player in the game, as it depends
 on the strategies of the agents.
 With this information, we estimated the expected value of the payoff per round
 for the remaining rounds for each choice.

 We assume that there are $M$ pieces ofbandit information in the player's repertoire at round $t$.
 We denote them as $(n_{m},s_{m},t_{m}),m \in \{1,\cdots,M\}$.  Here, $t_{m}$ is the round
 during which the player obtained the information. 
 When, the player obtains information
 from Innovate or obtains updated information from Exploit at $t'$, $t_{m}=t'$.
 If the player obtains information from Observe at $t'$, $t_{m}=t'-1$, as Observe returns the bandit
 information from the previous round, $t'-1$.

We denote the expected value of the payoff per round for exploiting 
bandit $n_{m}$ from $t$ to $T$ as $E_{m}(t)$. This quantity is estimated as
\begin{eqnarray}
&&E_{m}(t)=\mbox{E}(s)+\frac{1}{T-t+1}\sum_{t'=t}^{T}(s_{m}-\mbox{E}(s))(1-p_{c})^{t'-t_{m}}
\nonumber \\
&=&\mbox{E}(s)+\frac{(1-(1-p_{c})^{T-t+1})(s_{m}-\mbox{E}(s))(1-p_{c})^{t-t_{m}}}{p_{c}(T-t+1)}
\label{eq:Em}
\end{eqnarray}
where $(1-p_{c})^{t'-t_{m}}$ is the probability that 
the bandit information does not change
from $t_{m}$ until $t'$. During this period, the payoff is $s_{m}$.
If the bandit information changes until $t'$, the probability
 for it is $1-(1-p_{c})^{t'-t_{m}}$, and the expected payoff of 
the bandit is given by $\mbox{E}(s)$.
By summing these values and dividing by the number of rounds $T-t+1$, 
we obtain the above expression.

 We denote the expected payoff per round for Innovate as $I(t)$.
 For Innovate, a player does not receive any payoff. 
He only obtains bandit information,
 and the expected value of the
 payoff of the obtained bandit information is $\mbox{E}(s_{I})$. We estimate the
 expected value of the payoff by Innovate by assuming that the player continues
 to exploit the new bandit with the payoff $\mbox{E}(s_{I})$ from round $t+1$ to $T$ as
\begin{equation}
I(t)=\frac{T-t}{T-t+1}\mbox{E}(s)+
\frac{(1-(1-p_{c})^{T-t})(\mbox{E}(s_{I})-\mbox{E}(s))(1-p_{c})}{p_{c}(T-t+1)}.
\label{eq:I}
\end{equation}
As the player loses one round because of Innovate, the prefactor in front of $\mbox{E}(s)$
and the power of $(1-p_{c})$ are reduced to $(T-t)/(T-t+1)$ and $(T-t)$
as compared with those in eq.(\ref{eq:Em}). If $n_{I}=1$, $\mbox{E}(s_{I})=\mbox{E}(s)$,
the second term vanishes, and Innovate is almost worthless.
For cases in which all of the payoffs of the bandit information in one's repertoire are
zero or less than $\mbox{E}(s)$, it might be optimal to choose Innovate.
Otherwise, instead of losing one round and obtaining bandit information with a payoff $\mbox{E}(s)$,
 it is optimal to choose Exploit with the maximum expected payoff.
If $p_{c}$ is large, even if all the payoffs in one's repertoire is zero,
$(1-p_{c})^{t-t_{m}}$ can be negligibly small, and it is optimal to choose Exploit.
When $n_{I}>1$ and $p_{c}$ are not very large, Innovate might be optimal.

Likewise, we estimate the expected payoff per round for Observe, which we denote as $O(t)$.
For Observe, a player obtains bandit
information with a payoff $\overline{O}(t)$. The age of the information is one round older than the
information obtained by Innovate. We change $\mbox{E}(s_{I})$
 to $\overline{O}(t)$ in eq.(\ref{eq:I}). Accounting for the age of the
new bandit information, we estimate $O(t)$ as
\begin{equation}
O(t)=\frac{T-t}{T-t+1}\mbox{E}(s)+
\frac{(1-(1-p_{c})^{T-t})(\overline{O}(t)-\mbox{E}(s))(1-p_{c})^{2}}{p_{c}(T-t+1)}.
\label{eq:O}
\end{equation}
Comparing $I(t)$ and $O(t)$, which is more optimal depends on $p_{c}$ and
$\mbox{E}(s_{I})-\overline{O}(t)$. If $p_{c}$ is small and
$1-p_{c}\simeq 1$, the magnitude of the relationship between 
$\mbox{E}(s_{I})$ and $\overline{O}(t)$
determines which is more optimal.

The optimal strategy is to choose the action with maximum expected payoff during
every round $t\in \{-2,-1,\cdots,T=100\}$.
For example at $t=T$, the last round of the game, as $I(T)=O(T)=0$ holds, it is optimal to
choose Exploit for bandit $m$ with the maximum $E_{m}(T)$.
In the first three rounds where the player can choose
only Innovate or Observe,
if both $p_{c}$ and $n_{I}$ are small, $\mbox{E}(s_{I})<\overline{O}(t)$ usually holds.
 Observe is more optimal than Innovate in this case.
 The situation is the same in later rounds, and the optimal strategy is a combination
of Exploit and Observe.
Conversely, if both $p_{c}$ and $n_{I}$ are large, even if
$\overline{O}(t)\simeq \mbox{E}(s_{I})$, $(1-p_{c})<1$ and $I(t)>O(t)$ hold.

We estimate the expected payoff per round for several ``optimal'' strategies with a restriction
on the choice of learning. We consider three strategies, and an Exploit-only strategy as
a control strategy.

\begin{itemize}
\item I+O: The player can choose both Innovate and Observe when learning. In the first three rounds,
 Innovate is chosen. Then, the action with the highest expected payoff is chosen in the later rounds.

\item I: The player can choose Innovate for learning. The other conditions are the same as I+O.

\item O: The player can choose Observe for learning. The other conditions are the same as I+O.

\item EO: The player can choose Exploit with the maximum expected payoff after the
first three rounds.

\end{itemize}
The expected payoffs per round for these strategies are written as {\bf I+O}, {\bf I}, {\bf O},　
and {\bf EO}, respectively. We also denote the expected payoff per round for agent $i$ as {\bf P(i)}.
They are estimated by a Monte Carlo simulation. We have performed a simulation
of a game in which 120 agents and four players with above strategies participate
$10^{4}$ times. As we have explained in the
experimental procedure, the agents cannot observe the bandit information exploited by the player.
Only player can observe the bandit information of the agents.
As there is no interaction between the players,
we can estimate the expected payoffs of the four players simultaneously.
In the experiment, the player can choose Observe for the first three rounds. With the above strategies,
 the player can choose Innovate only for simplicity. The players and agents compete on equal terms.

\begin{figure}[tb]
\begin{center}
\includegraphics[width = 80mm]{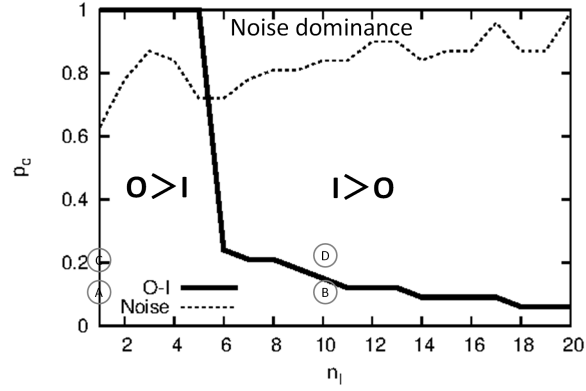}
\caption{Optimal learning in $(n_{I},p_{c})$.
The thick solid line shows the boundary between the region {\bf I}$>${\bf O} and the region {\bf O}$>${\bf I}.
The dotted line shows the boundary beyond which {\bf EO}$\simeq${\bf I+O}.}
\label{fig:I+O}
\end{center}
\end{figure}

 We summarize the results in Figure \ref{fig:I+O}.
 In $(n_{I},p_{c})$ plane, we show which strategy is more optimal, I or O.
 The thick solid line shows the boundary where {\bf I} = {\bf O}.
 In the lower-left region {\bf O}$>${\bf I} holds.
 As $n_{I}$ and $p_{c}$ are small, the relationship
 $\overline{O}(t)>\mbox{E}(s_{I})$ holds, and Observe becomes a optimal learning method.
 In the upper-right region, {\bf I} $>$ {\bf O} holds.
 $n_{I}$ is large, and $\mbox{E}(s_{I})$ is greater than or comparable to
 $\overline{O}(t)$. As $p_{c}$ is large, the one round delay for exploiting
 the bandit information obtained by Observe might be crucial.
 The thin dotted line shows the boundary beyond which
 {\bf I+O} is comparable with {\bf EO}. As $p_{c}$ is large, the player
 can obtain comparable payoffs by only exploiting a good bandit in his repertoire.
 There is neither an exploitation--extrapolation trade-off nor
 a social--asocial learning trade-off above the dotted line.
 It is a noise-dominant region.

 One can understand why there is no social--asocial learning trade-off
 in Rendell's tournament \cite{Ren:2010}. In the tournament, they set
 $n_{I}=1$ and $n_{O}\ge 1$. Here, $n_{O}$ is the amount of bandit information obtained by Observe.
 If $n_{I}=1$, as we have explained previously, $\mbox{E}(s_{I})=\mbox{E}(s)$ holds.
 If $p_{c}$ is small, $\overline{O}(t)$
 is usually greater than $\mbox{E}(s)$, as agents exploit the good bandit in their repertoire.
 Then, an agent can obtain good bandit
 information by Observe, and Observe becomes an optimal learning method.
 If $p_{c}$ is too large, instead of trying to obtain good information with Innovate, it is optimal
 to wait spontaneous changes in the bandit information in the repertoire.
 Exploiting a good bandit
 in one's repertoire (EO strategy) is enough, and no other strategy cannot exceed the performance
 of EO.

 In the region where {\bf O}$>${\bf I} and {\bf I+O}$>${\bf EO}, social learning is effective, and a
 swarm intelligence can emerge.
 We define the swarm intelligence effect as the increase in the performance compared to {\bf I}.
 In the next section, we estimate the swarm intelligence effect for human subjects.
 As for the choice of $(n_{I},p_{c})$, we have studied four cases A:(1,0.1), B:(10,0.1), C:(1,0.2),
 and D:(10,0.2). We show the positions for these choices in figure \ref{fig:I+O}.
 For cases A and C, {\bf O} $>>${\bf I}, and one expect to observe the swarm intelligence
 effect in human subjects.
 For cases B and  D,  where {\bf O}$\simeq${\bf I} and {\bf O}$<${\bf I},
 one does not expect to observe it.

 We make a comment about the definition of the swarm intelligence effect.
 For the estimation of {\bf I}, we assume that the player
 knows $p_{c}$, $\mbox{E}(s)$, and
 $\mbox{E}(s_{I})$ and can choose the best option among Exploit and Innovate.
 The player has to estimate this information from his actions in the real game.
 If the player cannot choose Observe, his performance cannot exceed {\bf I}.
 The definition of the swarm intelligence effect only provides a
 lower limit. Toyokawa \cite{Toy:2014} defined it as the surplus in performance
 compared to when the same player can only choose Innovate. Our definition has
 the advantage that it can be estimated easily without performing an 
 experiment.
 The same reasoning applies to {\bf I+O}. In this case, the player knows everything
 that is related with his decision making. {\bf I+O} provides an upper limit on
 the performance of the player in the game.

\section{Experimental results}
 In this section, we explain the experimental results.
 We estimate the swarm intelligence effect for human subjects.
 We perform a regression analysis of the performance of each subjects in each
 experimental environment.

\subsection{Swarm intelligence effect}
\begin{figure}[tb]
  \begin{center}
  \begin{tabular}{cc}
  \includegraphics[width=60mm]{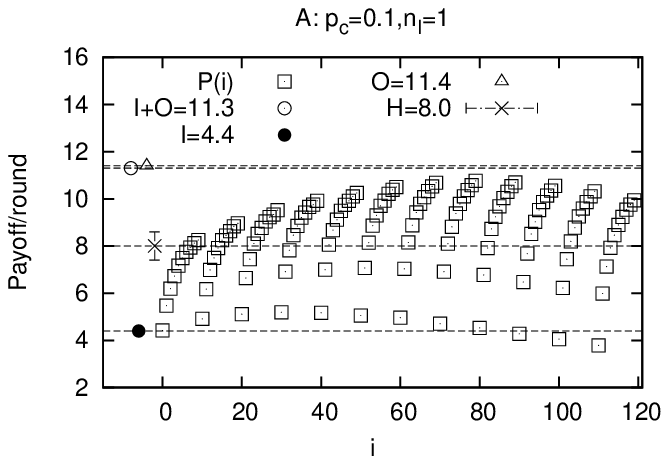} \
  \includegraphics[width=60mm]{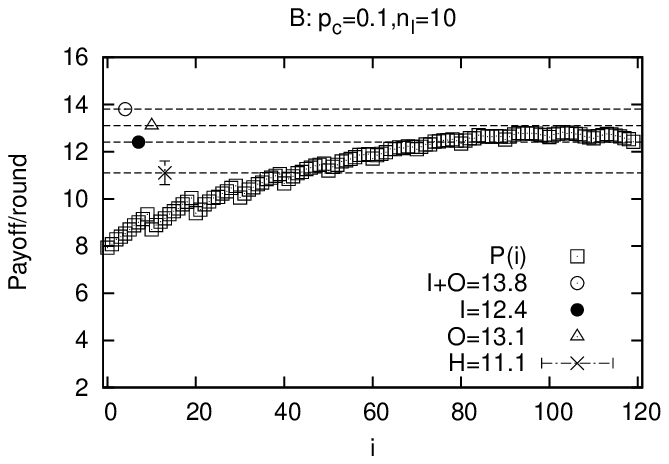} \\
  \includegraphics[width=60mm]{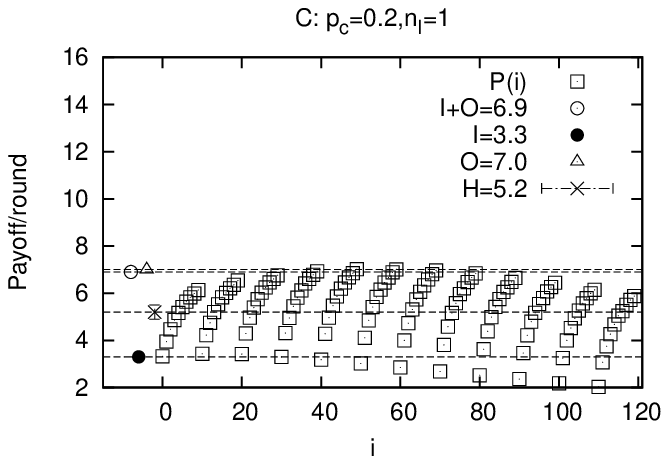} \
  \includegraphics[width=60mm]{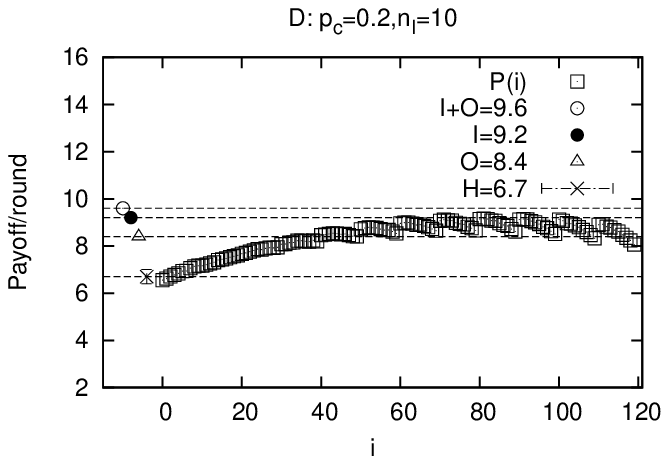}
  \end{tabular}
  \end{center}
  \caption{Plots of {\bf P(i)}($\Box$), {\bf I+O}($\circ$), {\bf I}($\bullet$), {\bf O}($\triangle$)  and {\bf H}($\times$). 
A:$(p_{c},n_{I})=(0.1,1)$, {\bf H}(54)$=8.0\pm 0.6$.
B:$(p_{c},n_{I})=(0.1,10)$, {\bf H}(65)$=11.1\pm 0.5$.
C:$(p_{c},n_{I})=(0.2,1)$, {\bf H}(54)$=5.2\pm 0.3$.
D:$(p_{c},n_{I})=(0.2,10)$, {\bf H}(52)$=6.7\pm 0.3$.
The number in each parentheses is the number of subjects in each case.
}
\label{fig:ABCD}
\end{figure}

 We calculated the total payoff of each subject for 100+3 rounds in each game environment.
 We divided the total payoff by 100 and obtained the average payoff per round.
 For each game environment, we estimated the average value
 of the average payoffs per round for approximately 56 subjects and denote it as {\bf H}.
 This represents the average performance of human subjects in each case.
 We compare {\bf H} with {\bf I+O}, {\bf I}, {\bf O}, and {\bf P(i)} for agent $i$.
 Figure \ref{fig:ABCD} show the results for cases A, B, C, and D.
 We explain the results of each case.

\begin{itemize}

\item[A:] Case A is in the region where where {\bf O}$>${\bf I}, and
Observe is optimal for learning.
As {\bf I+O}$\simeq${\bf O} and {\bf O} is much greater than {\bf I}, one can expect
 the swarm intelligence effect. In fact, {\bf H}, which is plotted with a chain line,
is higher than {\bf I}. For a fixed value of $c$,
{\bf P(i)} increases with $p_{obs}$. For the dependence of {\bf P(i)} on $c$
 for a fixed value of $p_{obs}$, there is a maximum for some $c$. For $p_{obs}=0.0$, {\bf P(i)} is maximum
at $c\sim 4$. For $p_{obs}=0.9$, {\bf P(i)} is maximum at $c\sim 8.5$. The agent can obtain
good bandit information by Observe, and they had better to adopt large $c$.

\item[B:] Case B is in the region where {\bf O}$>${\bf I}. However, it is
near the boundary for {\bf I} = {\bf O}, and the difference between {\bf O} and {\bf I} is small.
One cannot expect the swarm intelligence effect. In fact, {\bf H} is below {\bf I}.
As {\bf I}$\simeq ${\bf O}, subjects could not improve their performance by Observe.
One see {\bf I+O}$>${\bf O}, and the difference between {\bf I+O} and {\bf O} is small.
As $n_{I}=10$ is large, Innovate is frequently more optimal than Observe.
For example, if an agent finds that the payoff of good bandit information
changes to zero in round $t$ by Exploit, one can suspect that
Observe does not provide good bandit information. In particular, if the bandit is
good, and the payoff is high, one can assume that many agents also exploited the bandit.
Then Observe should provide bandit information with zero payoff in round $t+1$ with a high probability.

{\bf P(i)} is an increasing function of
$p_{obs}$ when $c$ is small. When $c$ is large, {\bf P(i)} does not depend on $c$ very much. When $c$ is small, agents can
 easily obtain bandit information whose payoff is greater than $c$. Then, the agent exploits
 the not so good bandit. On the other hand, if the agent obtains bandit information by Observe,
 he can obtain good bandit information, as the bandit's payoff exceeds the other agents' $c$.
 Observe is more optimal than Innovate when $c$ is small.
 However, when $c$ is large, the agent can obtain good bandit information by Innovate, as $n_{I}$ is large.
 By Observe, the agent can obtain good bandit information, and there is not a big difference in the
 performance of {\bf I} and {\bf O}. As a result, {\bf P(i)} does not depend very much on $p_{obs}$ when
 $c$ is large.

\item[C:] Case C is in the region where {\bf O}$>${\bf I}.
{\bf I+O}$\simeq$ {\bf O}, and {\bf O} is much greater than {\bf I}, as with case A.
One can observe the swarm intelligence effect because {\bf H} is greater than {\bf I}.
Because $p_{c}$ is large, the expected payoffs and average payoff of the subjects are lower than those for
case A.

\item[D:] Case D is in the region where {\bf I}$>${\bf O}, and Innovate is optimal for learning.
As {\bf I+O}$>${\bf I}, Observe is optimal in some cases. When $c$ is large, {\bf P(i)} is a
decreasing function of $p_{obs}$.
As both $p_{c}$ and $n_{I}$ are large, instead of obtaining good bandit information by Observe,
 Innovate succeeds in obtaining new and good bandit information. When $c$ is small,
 as in case B, the agent can obtain better bandit information by Observe than Innovate.
 One cannot observe the swarm intelligence effect, as in case B.
\end{itemize}

\subsection{Regression analysis of the performance of individual subjects}
 We perform a statistical analysis of the variation in the payoffs of the subjects
 in the four cases. We examined the factors that made strategies successful by using
 a linear multiple regression analysis. In Rendell's tournament, there were five predictors
 in the best-fit model for the performance of the strategies \cite{Ren:2010}.
 Among them, we considered three predictors: $r_{learn}$, the proportion of moves that
 involved learning of any kind;
 $r_{obs}$, the proportion of learning moves that were Observe; and $\Delta t_{learn}$, the average round
 between learning moves. Other predictors were the variance in the number of rounds to first use of Exploit and
 a qualitative predictor of whether or not the agent program estimates $p_{c}$.
 For the latter, we suppose that human subjects
 estimated $p_{c}$, or they could notice whether the frequency of the change in
 bandit information is high or low.
 For the former predictor, it is impossible to estimate it, as the subjects participated
 in the game at most once for each case.
 We do not include these two predictors in the regression model.
 We denote the average payoff per rounds for subject $j$ as $payoff(j)$.
The multiple linear regression model is written as
$payoff(j)=a_{0}+a_{1}\cdot r_{learn}(j)+a_{2}\cdot r_{obs}(j)+a_{3}\cdot \Delta t_{learn}(j)$.
 We select the model with maximum $\tilde{R}^{2}$. The results are summarized in Table \ref{tab:1}.

\begin{table*}[htb]
\begin{center}
\caption{Parameters of the linear multiple regression model predicting the average
payoff per round in each game environment.
From the second to fourth columns, the intercepts and regression coefficients
 for $r_{learn}$, $r_{obs}$, and $\Delta t_{learn}$ are shown.
n.s. for $p>0.05$,* for $p<0.05$,** for $p<10^{-2}$,*** for $p<10^{-3}$ and **** for $p<10^{-4}$.
}
  \begin{tabular}{|l|c|c|c|c|c|} \hline
    Case & Intercept & $r_{learn}$ & $r_{obs}$ & $\Delta t_{learn}$ & $\tilde{R}^{2}$ \\ \hline \hline
     A(53) & 10.0 (****)  & -11.0(**) & 3.3 $(p=0.16)$ & n.s. & 0.253 \\
     B(65) & 14.1 (****) &  -10.7 (*) & n.s. & n.s. &  0.076 \\
     C(54) & 8.37  (***)  & -8.2 (**) & 3.0 (*)  & -0.49 $(p=0.06)$ & 0.186 \\
     D(52) & 8.7 (****)  & -7.2 (**) & 1.4 $(p=0.23)$ & n.s. & 0.144 \\ \hline
     ALL(224) & 12.0 (****)   & -13.8 (****) & 1.3 $(p=0.15)$  & n.s. & 0.246 \\ \hline
  \end{tabular}
\label{tab:1}
\end{center}
\end{table*}

$r_{learn}$ had a negative effect on the performance of the subjects, as in Rendell's tournament.
This result suggests that it is suboptimal to invest too much time in learning, as one cannot obtain any
payoffs for learning. For $r_{obs}$, the results are not consistent with the results of
Rendell's tournament. There, the predictor had a strong positive effect, which reflected the fact that
 the best strategy was to almost exclusively choose Observe rather than Innovate.
 In our experiment, the predictor seems to have a positive effect for cases A, C, and D. For cases A and C,
   it is consistent with the results in the previous section because {\bf O}$>${\bf I}, and Observe is more
  optimal than Innovate. In case D, as {\bf H} is much less than both {\bf I} and {\bf O},
obtaining good bandit information from the agents by Observe might improve the performance.

\section{Conclusion}
In this paper, we attempt to clarify the optimal strategy in a two trade-offs environment.
Here, the two trade-offs are the trade-off of exploitation--exploration and that of social--asocial learning.
For this purpose, we have developed an interactive rMAB game, where a
player competes with multiple agents.
The player and agents choose an action from three options: Exploit a bandit,
 Innovate to obtain new bandit information, and Observe the bandit information exploited by other agents.
 The rMAB has two parameters, $p_{c}$ and $n_{I}$.
 $p_{c}$ is the probability for a change in the environment. $n_{I}$ is the scope of exploration
 for asocial learning.
 The agents have two parameters for their decision making, $p_{obs}$ and $c$.
 $p_{obs}$ is the probability for Observe when the agents learn, and
 $c$ is the threshold value for Exploit.

 We have estimated the average payoff of the optimal strategy with some restrictions on learning and
 complete knowledge about rMAB and the bandit information exploited during the previous round. We consider
 three types of optimal strategies, {\bf I+O}, {\bf I}, and {\bf O}, where both Innovate and Observe, Innovate, and
 Observe are available. In the $(n_{I},p_{c})$ plane, we have derived the strategy that is more optimal, either
 {\bf O} or {\bf I}. Furthermore, we have defined the swarm intelligence effect as the surplus of the performance
 of {\bf I}. The estimate of the swarm intelligence effect provides only a lower bound for it; however, the
 estimation is easy and objective. We also point out that the swarm intelligence effect can be observed
 in the region of the $(n_{I},p_{c})$ plane where {\bf O} is more optimal than {\bf I}. We have performed an experiment
 with 67 subjects and have gathered approximately 56 samples for the four cases of $(n_{I},p_{c})$.
 If $(n_{I},p_{c})$ are chosen in the region where {\bf O} is far more optimal that {\bf I},
 we have observed the swarm intelligence effect.
 If $(n_{I},p_{c})$ are chosen near the boundary of the two regions
 or in the region where {\bf I} is more optimal than {\bf O}, we did not observe the swarm intelligence effect.
 We have performed a regression analysis of the performance of each subject in each case.
 Only the proportion of learning is the effective factor in the four cases.
 In contrast, the proportion of the use of Observe for learning is not significant.

 As the agent's decision making algorithm is too simple,
 it is difficult to believe that the conditions for the emergence of the swarm
 intelligence in Figure \ref{fig:I+O} are general. In addition, the analysis of the human
 subjects is too superficial, as we only studied the correlation
 between the performance and some predictive factors. With these points in mind, we make
 three comments about future problems.

 The first one is a more elaborate and autonomous model of the decision
 making in an rMAB environment. The algorithm needs to estimate $p_{c}, \mbox{E}(s), \mbox{E}(s_{I})$, and
 $O(t)$ for round $t$ on the basis of the data that the agent has obtained through his choices.
 Then, the agent can choose the most optimal option during each round and maximize the
 expected total payoff on the basis of these estimates.
 This is an adaptive autonomous agent model.
 With this model, we can understand the decision making of humans in the rMAB game more deeply.
 It is impossible to understand human decision making completely with experimental data.
 On the basis of the model, we can detect the deviation in human decision making and propose a decision making
 model for a human that can be tested in other experiments.

The second one is the collective behavior of the above adaptive autonomous agents or humans.
It is necessary to clarify how the conditions for the emergence of the swarm intelligence
effect would change. In the case of a population of adaptive autonomous agents,
 they would estimate the optimal value of $p_{obs}$ for the environment $(n_{I},p_{c})$
 and collectively realize the optimal value. The optimal strategy should be
 neither {\bf I} nor {\bf O} but a mixed strategy of Innovate and Observe.
Then, the condition for the emergence of the swarm intelligence effect is that
the performance of {\bf I+O} is equal to that of {\bf I}. If the performance of the former is greater than that
of the latter for any $(n_{I},p_{c})$, the swarm intelligence effect can always emerge, except for the
noise-dominant region. After that, we can study the conditions with human subjects experimentally.
A human subject participates in the rMAB game as a player, as in this study, or many human players
participate in the game to compete with each other.
The target is how and when humans collectively solve the rMAB problem.

The third one is the design of an environment in which swarm intelligence works.
In this study, we choose the rMAB interactive game and study the conditions for the emergence
of the swarm intelligence effect for a player. However, there are many degrees of freedom
in the design of the game. For example, when an agent observes, there are many degrees of freedom
 regarding how bandit information is provided to the agent. In the present game environment, the probability that
a bandit exploited in the previous round is chosen is proportional to the number of agents who have exploited
 it. Instead, we can consider an environment in which the bandit information of the
 most exploited bandit is provided, the bandit information of the agents who are near the agent is
provided, or the player can choose a bandit by showing him the number of agents who have
exploited the bandit. We think these changes should
affect the choice and performance of the player. It was shown experimentally that by providing
subjective information about a bandit, the performance of the subjects diminished \cite{Toy:2014}.
We think that the interaction between the design of the environment and the decision making, performance, and
swarm intelligence effect should be a very important problem in the industrial usage of
the swarm intelligence effect.

\begin{acknowledgment}
We thank the referees for their useful comments and criticisms.
This work was supported by a Grant-in-Aid for Challenging
Exploratory Research 25610109.
\end{acknowledgment}


\begin{thebibliography}{99}
\bibitem{Aue:2002}Auer, P., Cesa-Bianchi, N. and Fisher, P.,
\newblock "Finite-time analysis of the multi-armed bandit problem,"
\newblock {\em Mach. Learn. 47}, pp. 235-256, 2002.

\bibitem{Ber:1985} Berry, D. and Fristedt, B., eds.,
\newblock {\em Bandit Problems: Sequential Allocation of Experiments},
\newblock Springer, Berlin, 1985.

\bibitem{Gal:2009} Galef, B. G.,
\newblock "Strategies for social learning: Testing predictions
from formal theory,"
\newblock {\em Adv. Stud. Behav. 39}, pp. 117-151, 2009.

\bibitem{Gir:2002} Giraldeau, L.-A.,Valone, T. J. and Templeton, J. J.,
\newblock "Potential disadvantages of using socially acquired information,"
\newblock {\em Philos. Trans. R. Soc. London Ser. B, 357}, pp. 1559-1566, 2002.

\bibitem{Gue:2014} Gueudr\'{e}, T.,Dobrinevski, A. and Bouchaud, J. P.,
\newblock "Explore or exploit? A generic model and an exactly solvable case,"
\newblock {\em Phys. Rev. Lett., 112}, pp. 050602-050606, 2014.

\bibitem{Kam:2003}Kameda, T. and Nakanishi, D.,
\newblock "Does social/cultural learning increase human adaptability? Roger's question revisited,"
\newblock {\em Evol. Hum. Behav., 24}, pp. 242-260, 2003.

\bibitem{Lai:1985}Lai, T. and Robbins, H.,
\newblock "Asymptotically efficient adaptative allocation rules,"
\newblock {\em Adv. Appl. Math., 6}, pp. 4-22, 1985.

\bibitem{Lal:2004}Laland, K. N.,
\newblock "Bandit problems: Sequential allocation of experiments,"
\newblock {\em Learn. Behav., 32}, pp. 4-14, 2004.

\bibitem{Pap:1999} Papadimitriou, C. H. and Tsitsiklis, J. N.,
\newblock "The complexity of optimal queueing network control,"
\newblock {\em Math. Oper. Res., 24}, pp. 293-305, 1999.

\bibitem{Ren:2010} Rendell, L. et al.,
\newblock "Why copy others? Insights from the social learning strategies tournament,"
\newblock {\em Science, 328}, pp. 208-213, 2010.

\bibitem{Sut:1998} Sutton, R. S. and Barto, A. G., eds.,
\newblock {\em Reinforcement Learning: An Introduction},
\newblock Cambridge, MIT Press, 1998.

\bibitem{Toy:2014} Toyokawa, W., Kim. H and Kameda, T.
\newblock "Human collective intelligence under dual exploration-exploitation dilemma,"
\newblock {\em PLoS ONE, 9}, p. e95789, 2014.

\bibitem{Whi:2012} White, J. M.,
\newblock {\em Bandit Algorithms for Website Optimization},
\newblock O'Reilly Media, 2012.

\end{thebibliography}
\end{document}